\def\year{2022}\relax
\documentclass[letterpaper]{article} 
\usepackage{aaai22}  
\usepackage{times}  
\usepackage{helvet}  
\usepackage{courier}  
\usepackage[hyphens]{url}  
\usepackage{graphicx} 
\usepackage{natbib}  
\usepackage{multirow}
\usepackage{caption} 
\usepackage[american]{babel}
\usepackage{microtype}
\DeclareCaptionStyle{ruled}{labelfont=normalfont,labelsep=colon,strut=off} 
\usepackage{amssymb}
\usepackage{algorithm}
\usepackage{algorithmic}

\usepackage{newfloat}
\usepackage{listings}
\floatstyle{ruled}
\newfloat{listing}{tb}{lst}{}
\floatname{listing}{Listing}
\title{MSP : Refine Boundary Segmentation via Multiscale Superpixel}
\author{
    Jie Zhu\textsuperscript{\rm 1}, Huabin Huang\textsuperscript{\rm 2}, Banghuai Li\textsuperscript{\rm 2}, Yong Liu\textsuperscript{\rm 3}, Leye Wang\textsuperscript{\rm 1}\thanks{corresponding author}
}
\affiliations{
	\textsuperscript{\rm 1}Peking University \quad \textsuperscript{\rm 2}Megvii Inc \quad
	\textsuperscript{\rm 3}Beihang University\\
	zhujie@stu.pku.edu.cn,
	\{huanghuabin,libanghuai\}@megvii.com,\\
	yongliu@buaa.edu.cn, leyewang@pku.edu.cn\\}

\begin{document}

\maketitle

\begin{abstract}
   In this paper, we propose a simple but effective message passing method to  improve the boundary quality for the semantic segmentation result. Inspired by the generated sharp edges of superpixel blocks, we employ superpixel to guide the information passing within feature map. Simultaneously, the sharp  boundaries of the blocks also restrict the message passing scope. Specifically, we average features that the superpixel block covers within feature map, and add the result back to each feature vector. Further, to obtain sharper edges and farther spatial dependence, we develop a multiscale superpixel module (MSP) by a cascade of different scales superpixel blocks. Our method can be served as a plug-and-play module and easily inserted into any segmentation network without introducing new parameters. Extensive experiments are conducted on three strong baselines, namely PSPNet, DeeplabV3, and DeepLabV3+, and four challenging scene parsing datasets including ADE20K, Cityscapes, PASCAL VOC, and PASCAL Context. The experimental results verify its effectiveness and generalizability.
\end{abstract}

\section{Introduction}
Semantic segmentation is a fundamental and challenging problem of computer vision,  whose goal is to assign a
semantic category to each pixel of the image. It is critical
for various tasks such as autonomous driving, image editing, 
and robot sensing. Recently, with the rapid development of deep learning, fully convolutional~\cite{long2015fully} based methods have been proposed to address the above task. These methods have achieved significant performance on various benchmarks~\cite{ADE20K, Cordts2016Cityscapes, everingham2015pascal, mottaghi2014role}. 
\begin{figure}[t]
	\begin{center}
		\includegraphics[width=1\linewidth]{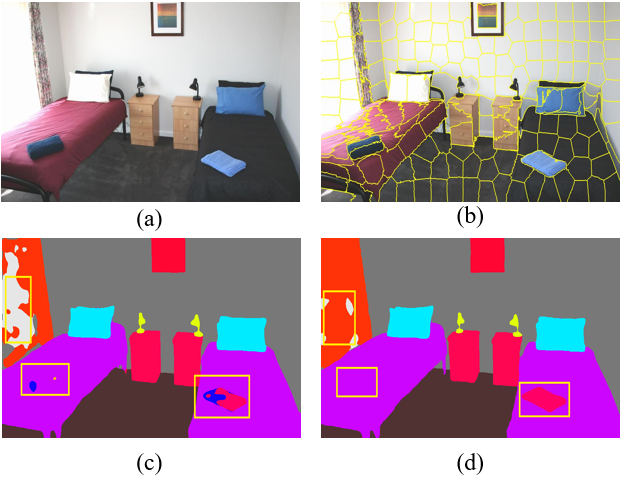}
	\end{center}
	\setlength{\belowcaptionskip}{0.2cm}
	\setlength{\abovecaptionskip}{0cm}
	\caption{\textbf{(a):} An image from ADE20K datasets \textbf{(b):} Superpixel that aggregates 
		similar pixels and groups them into blocks of different sizes and shapes according to low-level features such as the color, texture, and pixel location of the image. \textbf{(c):} Segmentation output from DeepLabV3+.   \textbf{(d):} Refined segmentation by our method.}
	\label{fig:sp}
\end{figure}
However, many researchers have also noticed that segmented boundary is of low quality to some extent. Therefore, many efforts are devoted to improving the performance of boundary segmentation. DenseCRF~\cite{krahenbuhl2012efficient}, as a classic method, uses color and position relationship of the 
original pixels of the image to refine the segmentation results containing poor boundary~\cite{chen2014semantic,chen2017deeplab,zheng2015conditional}.
However, DenseCRF usually serves as a post-processing module, which makes it not closely integrated with the CNN 
structure, and has a weak effect on optimizing the feature representation of edge points. Afterward, some works~\cite{wang2018non,huang2019ccnet} leverage attention mechanism of the high-level features  to construct more
reliable context information between the edge points and the internal points of the object. Some other works ~\cite{bertasius2016semantic, takikawa2019gated, yuan2020segfix} exploit as much boundary information as possible via deep neural networks. For example,  SegFix~\cite{yuan2020segfix} encodes the relative distance information of the boundary pixels with a boundary
map and a direction map, correcting the wrongly classified boundary pixels via internal points with high confidence. However, the accuracy of the weights or relationships obtained by these methods closely depends on 
the purity of the high-level features, which is what the edge feature points lack. The boundary features usually contain multiple object information due to the large receptive field of CNN. Therefore, the performance improvement brought by these methods is limited. 

Superpixels are an over-segmentation of an image that is formed by grouping image pixels based on the basic characteristics of the image, such as color, texture, and 
pixel position relationship. They provide a perceptually meaningful tessellation of image content, thereby reducing the number of image primitives.  Due to their representational
and computational efficiency, superpixels have turned into an accepted low/mid-level image representation. Many previous works~\cite{shu2013improving, yan2015object, zhu2014saliency} have employed this advantage and acquired pretty results. On the other hand, the generated pixel blocks usually contain sharp edges as shown in Fig.~\ref{fig:sp} (b), which is also worth exploring for downstream tasks like semantic segmentation.    

Thus, in this paper, we propose a simple but effective superpixel guided message passing method to correct wrongly segmented boundaries with the help of its sharp boundary. Further, inspired by the multiscale image pyramid, we design a multiscale superpixel information passing module (MSP), enabling multiple sharper edges and farther spatial dependence by a cascade of different scales superpixel blocks. These multiscale superpixel blocks are utilized to replace the high-level features to guide message passing within feature map. Simultaneously, the sharp  boundaries of the blocks also restrict the message passing scope, making neighboring boundary features acquire messages from different block sides. Finally, extensive evaluations of our multiscale superpixel algorithm on ADE20K, Cityscapes, PASCAL VOC, and PASCAL Context datasets are conducted to demonstrate its effectiveness and generalizability. A pair of segmentation visualization contrast can be found in Fig.~\ref{fig:sp} (c) and (d). The main contributions of this work are summarized as
follows: 

$\bullet$ We propose a simple but useful algorithm to refine semantic segmentation boundaries by superpixel because of its sharp edges and local consistency in a large area.

$\bullet$ A multiscale superpixel module is designed to obtain sharper edges and farther spatial dependence.

$\bullet$ Our method has obtained general improvement in semantic segmentation on three outstanding networks and four widely used scene parsing datasets.  

\section{Related Works}
\textbf{Semantic Segmentation.} Driven by powerful deep neural networks \cite{krizhevsky2012imagenet, simonyan2014very, szegedy2015going}, pixel-level prediction tasks like scene parsing and semantic segmentation achieve great progress inspired by replacing the fully-connected layer in classification with the convolution layer \cite{long2015fully}. To enlarge the receptive field of neural networks, several model variants are proposed. For example, GCN~\cite{peng2017large} adopts decoupling of large kernel convolution to gain a large receptive field for the feature map and capture long-range information. DeeplabV3 \cite{chen2017rethinking} extends ASPP with image-level feature to further capture global contexts. DeeplabV3+ \cite{chen2018encoder} adds a decoder upon DeeplabV3 to refine the segmentation results, especially along object boundaries.  The success of self-attention mechanism in natural language processing attracts lots of researchers' attention. DANet\cite{danet} applies both spatial and channel attention to gather information around the feature maps, which costs even more computation and memory than the Nonlocal\cite{nonlocal} method. EMANet\cite{li19} integrates  Expectation-Maximization (EM) algorithm to CNN to estimate attention map and reconstruct feature map while saving computing resources.   

\textbf{Superpixel.} Superpixel is pixels with similar characteristics that are grouped together to form a larger block. 
Since its introduction in 2003~\cite{ren2003learning}, there have been many pretty excellent algorithms\cite{achanta2012slic,weikersdorfer2013depth,van2012seeds} and mature evaluation metrics 
such as Undersegmentation Error. Moreover, publicly available 
superpixel algorithms have turned into standard tools in low-level vision. \cite{stutz2018superpixels} conducts fair analysis and evaluation of $28$ superpixel 
algorithms on $5$ datasets. Recently, in \cite{jampani2018superpixel}, neural network is 
applied to the generation of superpixel and great results are achieved. Superpixels have been applied in target 
detection 
\cite{shu2013improving,yan2015object}, 
semantic segmentation 
\cite{gould2008multi,sharma2014recursive,gadde2016superpixel}, saliency 
estimation 
\cite{he2015supercnn,perazzi2012saliency,yang2013saliency,zhu2014saliency}.  
\cite{yan2015object} converts object 
detection problem into superpixel 
labeling problem and 
conducts an energy function considering 
appearance, 
spatial context and numbers of labels.  
\cite{gadde2016superpixel} uses 
superpixels to change how information is 
stored in the 
higher level of a CNN. In   
\cite{he2015supercnn}, superpixels are 
taken as 
input and contextual information is 
recovered among 
superpixels, which enables large context 
to be involved 
in the analysis efficiently.

{\bf Refinement for Segmentation.} Previous work~\cite{zheng2015conditional,lin2017refinenet,chen2017deeplab} improved their segmentation results by DenseCRF~\cite{krahenbuhl2012efficient}. However, the low confidence score of the unary potential in boundary leads to a weak improvement of the object boundary segmentation, even with the help of pairwise potential. Recent works~\cite{acuna2019devil, chen2019learning} extended the conventional level-set scheme to deep
network for regularizing the boundaries of predicted segmentation map. Other studies~\cite{bertasius2016semantic, ding2019boundary, ke2018adaptive, takikawa2019gated, yuan2020segfix} also exploit the boundary information to improve the segmentation. These works aim at correct classification of edge pixels, by utilizing high-level features to directly or indirectly guide message passing for reliable boundary feature representations. For example,  SegFix~\cite{yuan2020segfix} encodes the relative distance information of the boundary pixels with a boundary
map and a direction map to correct the wrong boundary pixels. 

In this paper, we propose to refine segmentation via using superpixel blocks to guide the passing of information between features. Further, we design a plug-and-play multiscale superpixel module named MSP for sharper edges and longer dependency. Our method has been embedded in three famous semantic segmentation networks and evaluated on four challenging datasets. General gains brought by our method show its great potential.

\section{Approach}
In this section, we present our simple but effective method named multiscale superpixel module (MSP). Before that, we firstly give a detailed description of our single scale superpixel guided message passing model (SSP). Afterward, we introduce the multiscale superpixel model. Finally, we give an example of a combination with DeeplabV3+~\cite{chen2018encoder} that is considered as our baseline in most experiments and analyze the advantages of our proposed method.

\subsection{Superpixel Guided Message Passing} 
We use the superpixel segmentation algorithm to divide 
an image into hundreds of pixel blocks. These blocks define their respective scope of message passing. The $i$-th pixel block is denoted as $p_{i}$. In an image, all generated pixel blocks  belong to the set $P$. $K$ is the total number of elements in the set 
$P$. The entire information passing process consists of two steps: 1) computing averaged feature inside a superpixel; 2) adding it back to each pixel feature. Here, superpixel  plays a role of a mask. More specifically, for the $i$-th superpixel block $p_{i}$,  our approach averages the features inside the $p_{i}$ superpixel and adds the mean value $\bar{x_{i}}$ back to each feature vector covered by superpixel $p_{i}$. In 
order to obtain the mean feature map $\bar{X}$ of the 
entire feature map $X$, it is necessary to enumerate all pixel 
blocks from $1$ to $K$. Finally, $\bar{X}$ is weighted to the original $X$ to realize the message passing. The whole algorithm can also be formulated by the following formula:
\begin{equation}
\bar{X}=\sum_{i=1}^{K} {S(P_{i},X)\over{N(P_{i})}} \,,
\end{equation}
\begin{equation}
{S(P_{i},X)=\sum P_i\times X} \,,
\end{equation}
and 
\begin{equation}
{X^{*}=X+{\bf \alpha}\bar{X}} \,.
\end{equation}

$P_{i}$ is a binary map with the same size as $X$, 
and the value in the $p_{i}$ pixel block area is set to $1$ (otherwise  $0$). The function $S(, )$ sums feature value in $X$ along spacial dimension where the location is provided by $P_i$. The function $N(\cdot)$ calculates the area of 
the $pi$ pixel block. {\bf $\alpha$} is the weighting coefficient that is set to $0.1$ by default.

Thus, superpixel blocks, as low-level features, successfully guide the fusion of high-level information. This method is not only simple but also effective for the improvement of the boundary because the superpixel blocks with sharp boundaries allow adjacent edges from two or more objects to receive the information from respective inside  pixel blocks. This greatly boosts the discrimination of boundary features, especially for objects with obvious differences in characteristics. 

\begin{figure}[t]
	\begin{center}
		\includegraphics[width=0.9\linewidth]{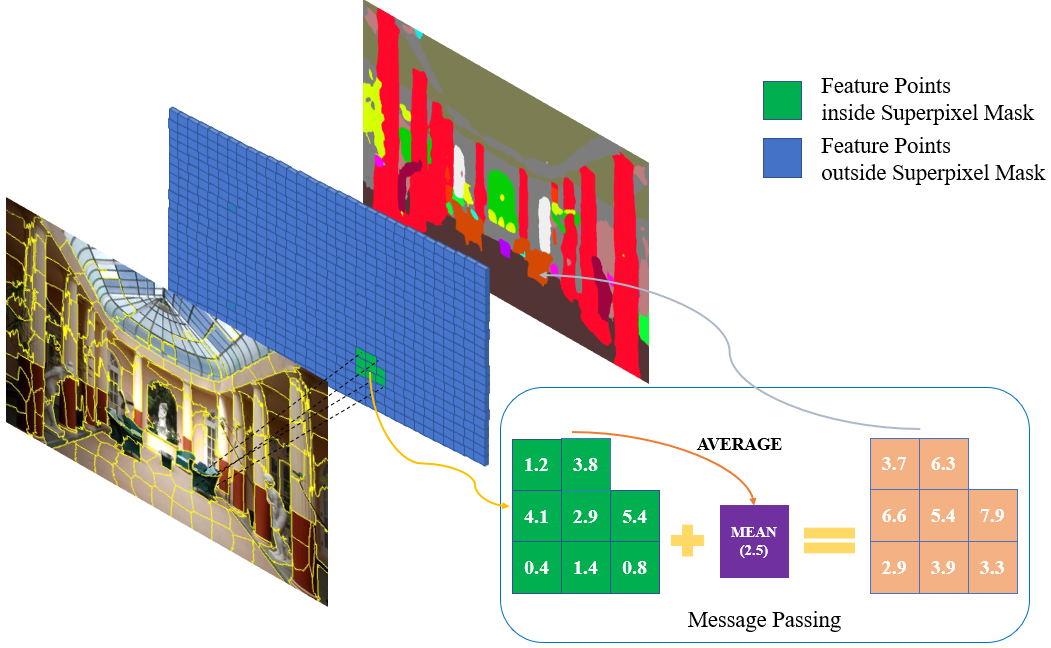}
	\end{center}
	\caption{Superpixel Guided Message Passing. A superpixel block is first mapped to a corresponding area in the feature map, and then the averaged feature in that area is regarded as a passing message and added back to the area.}
	\label{fig:message passing}
\end{figure}

In order to utilize the 
information of more features in a pixel block and avoid the excessive influence of a single feature vector, we average the features inside a pixel block and add it back to the original features with appropriate weight. In this way, each vector on the original feature map acquires the mean feature information within the 
pixel block where the vector is located. Note that this method introduces no convolution structure but still guarantees the backpropagation. The entire message passing process is presented in Fig. \ref{fig:message passing}.

In the experimental part, this method has achieved good results beyond the baseline, which proves its effectiveness.
\subsection{Multiscale Superpixel Module}
\begin{figure}
	\begin{center}
		\includegraphics[width=1\linewidth]{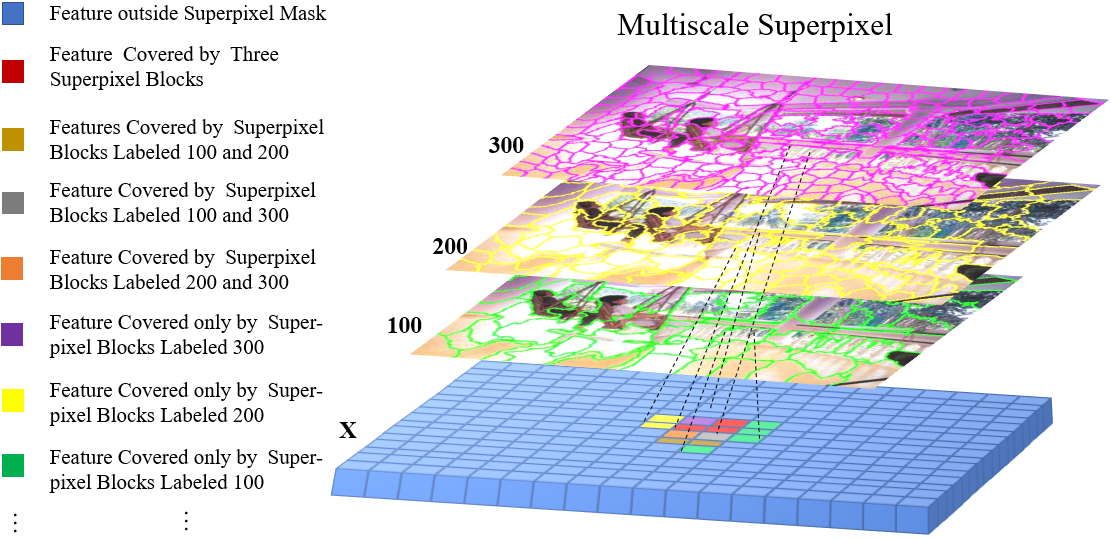}
	\end{center}
	\caption{Illustration on multiscale superpixel overlap. Due to different sizes of pixel blocks between different scales, some areas between pixel blocks of different scales may contain overlap to some extent.}
	\label{fig:multiscale}
\end{figure}

We consider that the single scale superpixel module (SSP) may lack diversity. And multiscale manner used to be a general solution that provides more details and broader context information. Therefore, we refer to the multiscale pyramid model and accordingly construct a multiscale superpixel model (MSP) in the process of information fusion. However, the key point where the model we design  
changes is not the size of the image but the 
number $\lambda$ of superpixel blocks in an image that is
divided. It is obvious that the shape of superpixel is irregular, which cannot be standardized and uniformly described. For the convenience of description, we define {\bf $\lambda$} as the scale of superpixels. It is worth noticing that the larger the $\lambda$ is, the greater the number of superpixel blocks in an image is, and the smaller the area of each pixel block is. In other words, a pixel block in an image may have a certain overlap with several pixel blocks on the other superpixel scales of that image, or it may be a part of a pixel block on another smaller scale as shown in Fig.~\ref{fig:multiscale}. As a result, the multiscale model makes a wider range of cross-fusion of features available.

In terms of the specific implementation, the multiscale model is formed by cascade single scale models. This makes 
the information passing of every single scale 
independent from each other, thereby avoiding the problem of message confusion between multiple scales in a parallel  fashion. In practice, the cascade fashion is easy to implement. Moreover, when cascade, the  order of superpixel scale is from a small scale to a large scale, such as 100, 200, and 300. In other words, we first conduct the message passing in a large superpixel area, then followed by a smaller one. Such a sequence guarantees that information holds longer dependence as small scale superpxiel often has a certain overlap with large scale.

In order to present our multiscale model more clearly and intuitively, we attempt to use a simple formula to illustrate it. Firstly, we define the single scale model as the function $F(X,\lambda,I)$, and $X^{*}$ is its output. So we can obtain the following formula:
\begin{equation}
	{X^{*}=F(X,\lambda,I)} \,,
\end{equation}
where $X$ is the original input feature, $\lambda$ (superpixel scale) is the number of superpixel blocks in a picture, and $I$ is a raw RGB image that is utilized to generate superpixel.

Then we present the formulas for a multiscale model based on the cascade of single scale models:
\begin{equation}
{X^{*}_{0}=F(X,\lambda_{0},I)}\,,
\end{equation}
\begin{equation}
{X^{*}_{1}=F(X^{*}_{0},\lambda_{1},I)}\,,
\end{equation}
\begin{equation}
\label{fo}
{X^{*}_{n-1}=F(X^{*}_{n-2},\lambda_{n-1},I)}\,.
\end{equation}

As is indicated in the above formulas, the output of the previous stage is used as the input for the next stage.

Finally, it is necessary to clarify the meaning of the parameters in fomula~(\ref{fo}) as well as an important rule. $n$ is the number of cascade single scale models. $\lambda_{i}$ is the number of superpixel blocks at the $i$-th scale. As we defined before, the scale here is not the size but the number of superpixel blocks. And keep $\lambda_{i}>\lambda_{i-1}$ in mind.

The performance of 
the multiscale superpixel model exceeds that of a single scale in the experiments of ADE20K below, which proves the effectiveness of the multiscale model and further demonstrates the superiority of superpixel in guiding the message passing of high-level features.

\subsection{Network Architecture}
\begin{figure*}
	\begin{center}
		\includegraphics[width=0.9\linewidth]{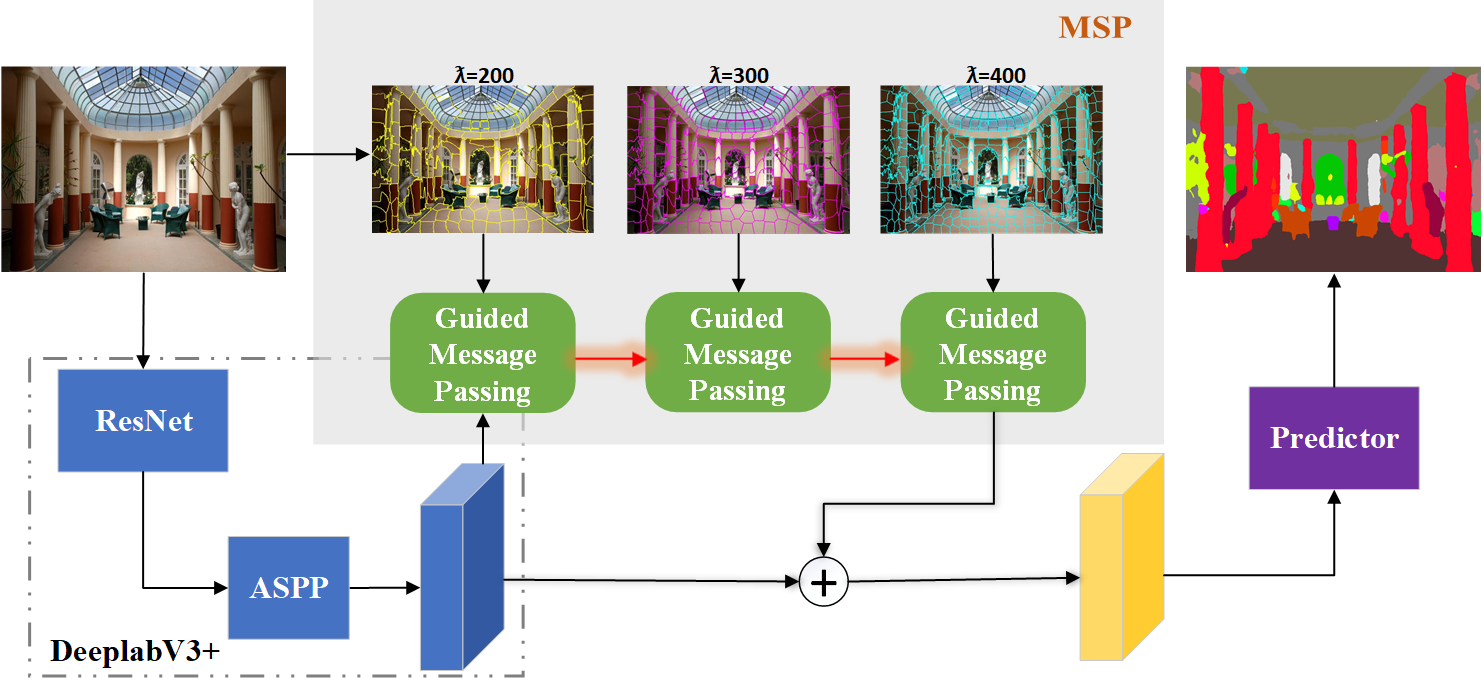}
	\end{center}
	\caption{An overview of our network based on DeeplabV3+.}
	\label{fig:short}
\end{figure*}

In this paper, We use DeeplabV3+ \cite{chen2018encoder} as the baseline of our experiments. 
The information passing of superpixel is carried out after the depthwise separable convolution layers of DeeplabV3+, which is before the classifier. The overall structure is shown in Fig. \ref{fig:short}, where the superpixel model we design is plug-and-play and can be easily  embedded in the network for end-to-end training and testing. Besides, as is described above, our method is simple to implement but the performances in the next section show its effectiveness and generalizability thanks to the sharp edge of superpixel blocks.
Finally, since no additional convolution structures are introduced, the parameters of the network model are not increased. 
\section{Experiment}
To evaluate the proposed method, we conduct extensive experiments on three outstanding neural networks and four widely used sense parsing datasets.  In this section, we 
first introduce implementation 
details followed by the comparisons with our baselines. Then we perform ablation 
studies to verify the 
superiority of the proposed method 
on the ADE20K dataset. 
Besides, we compare our method with DenseCRF and SegFix both on mIoU and F-score. Finally, we report our results on the Cityscapes dataset, PASCAL VOC dataset, and PASCAL Context dataset.

\subsection{Implementation Details}
All our experiments are based on MMSegmentation~\cite{mmseg2020}. We use
ResNet~\cite{7780459} (pretrained on ImageNet~\cite{russakovsky2015imagenet}) as our backbone. The output stride of the backbone is set to 8.
Following prior works~\cite{chen2017rethinking,zhao2017pyramid}, 
we employ a poly learning rate 
strategy where the initial learning
rate is multiplied by ${(1 -iter/total iter)}^{0.9}$ after each iteration, which is set to $80000$ as the maximum number of iterations in all experiments. The initial 
learning rate is set to be 0.01 
for all datasets. Momentum and weight decay coefficients are set
to 0.9 and 0.0005, respectively. For data augmentation, we
apply the common scale (0.5 to 2.0), cropping and flipping
of the image to augment the 
training data. The synchronized 
batch normalization is adopted in all experiments, together 
with the multi-grid~\cite{chen2017rethinking}.

Input size for 
ADE20K dataset is set to $512\times512$, while input size for 
Cityscapes dataset is set to $769\times769$. For PASCAL VOC and PASCAL Context, the input size is set to $512\times512$ and $480\times480$ respectively.  The batch size on ADE20K, PASCAL VOC, and PASCAL Context is set to 16 and Cityscapes is set to 8 due to the limited calculation resource. We train 40K iterations on PASCAL VOC and 80K iterations on ADE20K,  Cityscapes, PASCAL Context.
\subsection{Comparisons with Baselines on ADE20K}
In order to prove the 
effectiveness of the proposed method, we 
compare with DeeplabV3+~\cite{chen2018encoder} on the 
validation set of ADE20K. We report the mIoU of each network structure on 
different backbones in Tab.~\ref{Tab 1}. 
It is shown that the network structures equipped with our superpixel-based method have achieved excellent performances  compared with the original ones. More specifically, our single scale method based on DeeplabV3+ with backbone ResNet101 achieves 45.47\% in mIoU, and outperforms the original one by 0.56\%.  
The multiscale solution with ResNet-50 and ResNet-101 achieves 43.93\%  and 45.81\% respectively in mIoU, and outperforms the single scale solution by 0.61\% and 0.34\% respectively. A multiscale manner can 
compensate for the omission of 
information caused by a single scale, 
and this fashion can simultaneously capture longer and more effective dependency as well. Some visualization results compared with baseline are shown in Fig.~\ref{fig:result}.

\begin{table}[t]
	\begin{center}
		\begin{tabular}{c|c|c c|c}
			\hline
			Method & Backbone & SSP & MSP & mIoU(\%) \\
			\hline\hline
			DeeplabV3+ & ResNet50 &  & & 42.91\\
			DeeplabV3+ & ResNet50 &  \checkmark & &43.32\\						
			DeeplabV3+ & ResNet50 &  & \checkmark &{\bf 43.93}\\
			\hline	
			DeeplabV3+ & ResNet101 & & & 44.91\\
			DeeplabV3+ & ResNet101 & \checkmark & &45.47\\
			DeeplabV3+ & ResNet101 & & \checkmark &{\bf 45.81}\\
			\hline
		\end{tabular}
	\end{center}
	\setlength{\abovecaptionskip}{0.2cm}
	\caption{{\bf SSP:} Single scale superpixel Module. {\bf MSP:} Multiscale superpixel Module. The multiscale method works better than the single scale approach, and shows great potential. We believe that the reason is that the multiscale method contains more useful information and larger spatial dependencies. In the single scale model, $\lambda$ is set to 200, while in the multiscale model, $\lambda$ is set to 200, 300, and 400 respectively.}
	\label{Tab 2}
\end{table}

\subsection{Ablation Studies on ADE20K}
The ADE20K dataset is one of the most challenging benchmarks, which contains 150 classes and a variety
of scenes with 1,038 image-level labels. We follow the official protocol to split the whole dataset. Like most previous
works, we use the mean of Intersection over Union (mIoU) for evaluation. Single scale images are adopted as input for testing by default if not specified. For ablation experiments, we
adopt ResNet-50 and ResNet-101 as our backbones. 
\begin{figure*}[t]
	\begin{center}
		\includegraphics[width=0.95\linewidth]{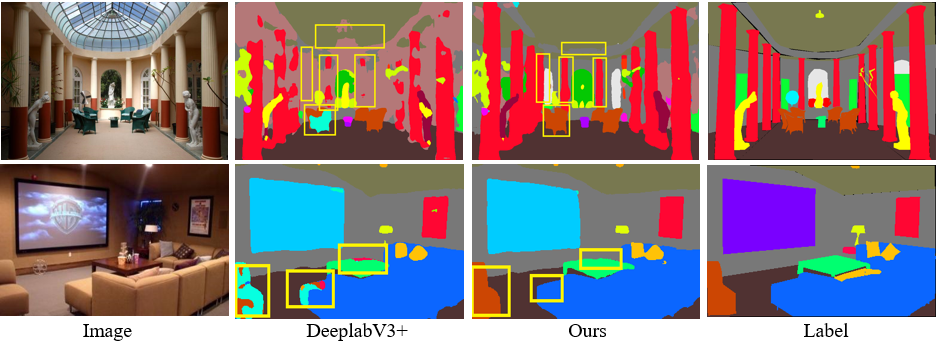}
	\end{center}
	\setlength{\abovecaptionskip}{0cm}
	\caption{Qualitative comparisons between our method and baseline on ADE20K validation set. It can be seen that our method can better segment objects with consistent textures or colors, such as the pillars and sofas in the first row. Moreover, it can also smooth the edges better, such as the sofa and coffee table in the second row. These all result from the local similarity of the low-level features of superpixels and the sharp edges between pixel blocks.}
	\label{fig:result}
\end{figure*} 

\subsubsection{Different Superpixel Algorithms} To explore how different algorithms affect the segmentation performance, we have conducted an ablation study on three different kinds of algorithms. They are density-based Quick Shift (QS)~\cite{vedaldi2008quick},  clustering-based SLIC~\cite{achanta2012slic}, and CNN-based SSN~\cite{jampani2018superpixel}. We adopt single scale superpixel guided message passing method and set $\lambda$ to 200 for all the experiments. DeeplabV3+ with backbone ResNet-50 is served as our baseline. The result is reported in Tab.~\ref{Tab al}. 
Though SSN achieves the best performance, it needs extra training time. On the other hand, SLIC, as an unsupervised clustering-based superpixel method, also performs well. Thus, we use SLIC for a fast implementation. 
\setlength{\tabcolsep}{0.5cm}{\begin{table}[t]
	\begin{center}
		\begin{tabular}{c c c|c}
			\hline
			QS & SLIC & SSN & mIoU(\%) \\
			\hline\hline
			 &  &  & 42.91\\
			 \hline
		\checkmark &  &  & 43.19\\
			 & \checkmark & & 43.32\\
			 &  & \checkmark & 43.47\\
			\hline	
		\end{tabular}
	\end{center}
	\setlength{\abovecaptionskip}{0.2cm}
	\caption{Comparisons with different superpixel algorithms.}
	\label{Tab al}
\end{table}
\subsubsection{Different Numbers of Superpixel and Scales} In order to explore the internal relationship between segmentation performance and superpixel property,  we have conducted abundant experiments on different numbers of scales and combinations of different superpixel numbers. As the two ablation studies are deeply correlative, we integrate them together for better comparisons. The results are reported in Tab.~\ref{Tab scale}. As is indicated in this table, it seems that multiscale $\lambda$ set to 200, 300, and 400 performs the best. Obviously, it achieves a  beautiful balance between large dependency and superpixel quality. 

\setlength{\tabcolsep}{0.25cm}{\begin{table}[h]
		\begin{center}
			\begin{tabular}{c|c c c c c|c}
				\hline
				Scale & 100 & 200& 300 & 400 & 500 &mIoU(\%)  \\ 
				\hline\hline
				  0 & & & & && 42.91\\
				\hline 
				\multirow{4}{*}{1}
				 & \checkmark & & & &&42.84 \\
				 &  & \checkmark & & &&43.32\\
				  &  &  & \checkmark & &&43.39\\
				  &  &  &  & \checkmark &&43.52\\
				  &  &  &  &  &\checkmark&42.59\\
				\hline
				\multirow{3}{*}{2}
				& \checkmark &\checkmark & & &&43.12 \\
				&  & \checkmark & \checkmark & &&43.48\\
				&  &  & \checkmark & \checkmark&&43.74 \\
				\hline
				\multirow{2}{*}{3}
				& \checkmark & \checkmark & \checkmark& &&43.36 \\
				&  & \checkmark &\checkmark & \checkmark && {\bf 43.93}\\
				\hline
				4 & \checkmark &\checkmark & \checkmark& \checkmark &&43.56\\
				\hline
			\end{tabular}
		\end{center}
		\setlength{\belowcaptionskip}{-0.1cm}
		\setlength{\abovecaptionskip}{0.2cm}	
		\caption{Comparisons on the ADE20K val set with DeepLabV3+ (ResNet50) as baseline. Since the damage is up to {\bf 0.32\%} while $\lambda$ is set to $500$, we no longer take $\lambda=500$ into account for multiscale.} 
		\label{Tab scale}
\end{table}}
\setlength{\tabcolsep}{0.2cm}{\begin{table}[t]
		\begin{center}
			\begin{tabular}{c c c|c c}
				\hline
				DenseCRF & SegFix & MSP & mIoU(\%) & F-score \\ 
				\hline\hline
				&  & & 44.91& 20.23\\
				\checkmark &  & & 45.43 & 21.86 \\
				&  \checkmark & & 45.62 & 22.19\\
				&  & \checkmark &{\bf 46.61} & 22.34\\
				\hline
				\checkmark & & \checkmark& 46.62 & 23.26\\
				& \checkmark & \checkmark&{\bf 47.05}& 24.01\\
				\checkmark & \checkmark & \checkmark & 47.00 & 24.65\\
				\hline
			\end{tabular}
		\end{center}
		\setlength{\abovecaptionskip}{0.2cm}
		\caption{Comparisons with DenseCRF and SegFix on ADE20K validation set. We take Deeplabv3+ (ResNet101) as the baseline. The post-processing method of DenseCRF and SegFix brings slightly lower gains than our multiscale superpixel solution. When the multiscale superpixel method is combined with SegFix, the mIoU on the ADE20K validation set reaches {\bf 47.05\%}.}
		\label{Tab 3}
\end{table}}
\subsection{Comparisons with DenseCRF and SegFix}
The sharp edges of the superpixel blocks allow the adjacent boundary features from different sides to receive the information from 
corresponding inside superpixel blocks when the message is passing. In other words, superpixel greatly 
enhances the separability of the boundary because of its own sharp boundary. 
In this section, we compare our method with some edge optimization algorithms. Here we mainly 
choose DenseCRF~\cite{krahenbuhl2012efficient} and SegFix~\cite{yuan2020segfix}. We keep the 
same setting of DenseCRF with 
Deeplab~\cite{chen2014semantic} and fine-tune  its parameters for better performance. As for SegFix, we follow the training strategy in  ~\cite{yuan2020segfix}, and train SegFix 80000 iterations on the ADE20K 
training set. Then for fair comparisons, the two post-processing algorithms are applied to refine the prediction of DeeplabV3+ on the ADE20K validation set, in which our method is embedded. In addition, we also superimpose the two 
algorithms and our method to further 
improve the segmentation on the 
ADE20K validation set.  

The result is reported in Tab. \ref{Tab 3}.
As is indicated in the table, our solution reaches the best performance on both mIoU and F-score among all the algorithms.
And it can be seen that DenseCRF has brought a limited gain in mIoU and has a weak effect on the improvement of the object boundaries. Moreover, our experiments show that our method is also
complementary with the SegFix.

\subsection{Generalizability on Other Methods}
To verify the effectiveness and generalizability of our proposed MSP, we have conducted extensive experiments based on two different baselines, namely PSPNet~\cite{zhao2017pyramid} and DeeplabV3~\cite{chen2017rethinking} on the validation set of ADE20K. The results are reported in Tab.~\ref{Tab 1} and Tab.~\ref{Tab deep}. As is indicated, our method can bring a relatively large gain to the two networks. Specifically, our method based on PSPNet with backbone ResNet-101 achieves 44.41\% in mIoU, and outperforms the original one by 0.84\%. Our method based on DeeplabV3 with backbone ResNet-101 achieves 44.89\% in mIoU, and outperforms the original one by 0.81\%. 
\setlength{\tabcolsep}{0.4cm}{\begin{table}
		\begin{center}
			\begin{tabular}{c|c|c|c}
				\hline
				Method & Backbone & MSP & mIoU(\%) \\
				\hline\hline
				PSPNet & ResNet50 &  & 41.23\\
				PSPNet & ResNet50 & \checkmark &{\bf 41.99}\\
				\hline			
				PSPNet & ResNet101 &  & 43.57\\
				PSPNet & ResNet101 & \checkmark &{\bf 44.41}\\
				\hline			
			\end{tabular}
		\end{center}
		\setlength{\belowcaptionskip}{-0.2cm}
		\setlength{\abovecaptionskip}{0.1cm}
		\caption{Comparisons with PSPNet on ADE20K dataset.}
		\label{Tab 1}
\end{table}}
\setlength{\tabcolsep}{0.36cm}{\begin{table}
		\begin{center}
			\begin{tabular}{c|c|c|c}
				\hline
				Method & Backbone & MSP & mIoU(\%) \\
				\hline\hline
				DeeplabV3 & ResNet50 &  & 42.42\\
				DeeplabV3 & ResNet50 & \checkmark &{\bf 43.16}\\
				\hline			
				DeeplabV3 & ResNet101 &  & 44.08\\
				DeeplabV3 & ResNet101 & \checkmark &{\bf 44.89}\\
				\hline			
			\end{tabular}
		\end{center}
		\setlength{\belowcaptionskip}{-0.2cm}
		\setlength{\abovecaptionskip}{0.1cm}
		\caption{Comparisons with DeeplabV3 on ADE20K dataset.}
		\label{Tab deep}
\end{table}}
\subsection{Generalizability on Other Datasets}

To further verify the effectiveness and generalizability of our proposed MSP, we have conducted extensive experiments on three another datasets, namely Cityscapes, PASCAL VOC, and PASCAL Context. 
\subsubsection{Cityscapes} 
Cityscapes is another popular dataset for scene parsing, which contains totally 19 classes. It consists of
5K high-quality pixel-annotated images collected from 50
cities in different seasons, all of which are with $1024\times2048$
pixels. In this data set, the training set contains 2975 finely annotated pictures, the validation set contains 500 pictures, and the test set contains 1525 pictures. 
\setlength{\tabcolsep}{0.35cm}{\begin{table}[h]
		\begin{center}
			\begin{tabular}{c|c|c|c}
				\hline
				Method & Backbone & MSP & mIoU(\%)  \\ 
				\hline\hline
				DeeplabV3+ & ResNet50 & & 79.24\\
				DeeplabV3+ & ResNet50 & \checkmark & {\bf 79.79}  \\
				\hline
				DeeplabV3+ & ResNet101 & &79.93 \\
				DeeplabV3+ & ResNet101 &\checkmark &{\bf 80.49}\\
				\hline
			\end{tabular}
		\end{center}
		\setlength{\abovecaptionskip}{0.2cm}
		\setlength{\belowcaptionskip}{-0.3cm}
		\caption{Comparisons on the Cityscapes validation set with baseline DeeplabV3+. All experiments in the table adopt single scale image as input for network. And since the raw image is 1024 × 2048, in order to maintain the quality of the superpixel block, we  adjust $\lambda$ to 100 and 200 naturally while in MSP mode. {\bf MSP :} Multiscale Superpixel Module.}
		\label{Tab 5}
\end{table}}

For the sake of a fair comparison, we adopt ResNet-50 and ResNet-101 as our backbones respectively.  Taking DeeplabV3+ as the baseline, we use 2975 finely annotated images in the Cityscapes dataset for training, and 500 images in the validation set and 1525 images in test set for evaluation with single scale input. The results can be found in Tab. \ref{Tab 5} and Tab.~\ref{Tab 7} respectively. It is obvious that the proposed approach outperforms DeeplabV3+ in both val and test sets. More specifically, We achieve {\bf 80.49\%} and {\bf 80.14\%} in mIoU with ResNet-101 as backbone on the validation and test set, outperforming the baselines by {\bf 0.56\%} and {\bf 0.92\%}, which further demonstrates the effectiveness of our method.
\setlength{\tabcolsep}{0.35cm}{\begin{table}[h]
		\begin{center}
			\begin{tabular}{c|c|c|c}
				\hline
				Method & Backbone & MSP & mIoU(\%)  \\ 
				\hline\hline
				DeeplabV3+ & ResNet50 & & 78.23\\
				DeeplabV3+ & ResNet50 & \checkmark & {\bf 78.98}  \\
				\hline
				DeeplabV3+ & ResNet101 & &79.22 \\
				DeeplabV3+ & ResNet101 &\checkmark &{\bf 80.14}\\
				\hline
			\end{tabular}
		\end{center}
		\setlength{\abovecaptionskip}{0.2cm}
		\setlength{\belowcaptionskip}{-0.3cm}
		\caption{Comparisons on the Cityscapes test set with baseline DeeplabV3+. We keep the same setting as cityscapes val set.}
		\label{Tab 7}
\end{table}} 
\setlength{\tabcolsep}{0.35cm}{\begin{table}[h]
		\begin{center}
			\begin{tabular}{c|c|c|c}
				\hline
				Method & Backbone & MSP & mIoU(\%)  \\ 
				\hline\hline
				DeeplabV3+ & ResNet50 & & 76.81\\
				DeeplabV3+ & ResNet50 & \checkmark & {\bf 77.64}  \\
				\hline
				DeeplabV3+ & ResNet101 & &78.62 \\
				DeeplabV3+ & ResNet101 &\checkmark &{\bf 79.49}\\
				\hline
			\end{tabular}
		\end{center}
		\setlength{\abovecaptionskip}{0.2cm}
		\caption{Comparisons on the PASCAL VOC set with baseline DeeplabV3+. We set $\lambda$ to 100 and 200 while in MSP mode. Single scale images are also adopted as input for network in all experiments. }
		\label{Tab 8}
\end{table}}
\subsubsection{PASCAL VOC} 
Experiments on the PASCAL VOC dataset are conducted based on DeeplabV3+ with ResNet-101 and ResNet-50 as the backbone, respectively. Quantitative results of PASCAL VOC are shown in Tab.~\ref{Tab 8}. Our method has outperformed baseline remarkably and brought {\bf 0.83 \%} (ResNet-50) and {\bf 0.87\%} (ResNet-101) gains in mIoU. It seems that our method is generally beneficial.
\subsubsection{PASCAL Context}
We have conducted experiments on the PASCAL Context dataset as well. In the experiments, we set $\lambda$ to 100 and 200 while in MSP mode following PASCAL VOC. When it comes to evaluation, we adopt single scale images as input for network. Comparisons with baseline are shown in Tab.~\ref{Tab all}. As is indicated in the table, our approach achieves {\bf 48.11\%} in mIoU and outperforms DeeplabV3+ by {\bf 0.84\%} with ResNet-101 as the backbone.
\setlength{\tabcolsep}{0.35cm}{\begin{table}[h]
		\begin{center}
			\begin{tabular}{c|c|c|c}
				\hline
				Method & Backbone & MSP & mIoU(\%)  \\ 
				\hline\hline
				DeeplabV3+ & ResNet101 & &47.27 \\
				DeeplabV3+ & ResNet101 &\checkmark &{\bf 48.11}\\
				\hline
			\end{tabular}
		\end{center}
		\setlength{\abovecaptionskip}{0.2cm}
		\setlength{\belowcaptionskip}{-0.3cm}
		\caption{Comparisons on the PASCAL Context set.}
		\label{Tab all}
\end{table}} 

\section{Conclusion}
In this paper, we propose a simple but effective message passing method for the use of superpixel that contains sharp boundary in semantic segmentation, which has brought general gains to our baselines on ADE20K, Cityscapes, PASCAL VOC, and PASCAL Context datasets. However, we just make a small step of exploration in how to use superpixel appropriately.
In our opinion, mature superpixel algorithms are of great help to semantic segmentation and even various aspects of computer vision tasks, such as instance segmentation, object detection, saliency estimation, and so on. We will also attempt to dive deep into the follow-up 
research. 

{\small
\bibliography{aaai22}
}

\end{document}